\documentclass{article}
\usepackage{arxiv}
\usepackage[T1]{fontenc}
\usepackage[utf8]{inputenc}
\usepackage{times}
\usepackage{amsmath,amssymb,amsfonts}
\usepackage{booktabs,tabularx}
\usepackage{graphicx}
\usepackage[table]{xcolor}
\usepackage{tcolorbox}
\tcbuselibrary{listingsutf8,breakable}
\usepackage{natbib}
\usepackage{url}
\usepackage{hyperref}
\usepackage{microtype}
\hypersetup{colorlinks=true,citecolor=blue,linkcolor=blue,urlcolor=blue}
\title{Document-Level Text Simplification in Estonian Using Large Language Models}
\author{Meeri-Ly Muru, Eduard Barbu \\
National Library of Estonia \\
Institute of Computer Science, University of Tartu \\
Tartu, Estonia \\
\texttt{meeri-ly.muru@rara.ee, eduard.barbu@ut.ee}}
\date{}
\begin{document}
\maketitle
\begin{abstract}

Document-level text simplification involves transformations that go beyond sentence-internal edits, addressing discourse coherence, anaphora resolution, and cross-paragraph consistency. Despite advances in sentence-level simplification for high-resource languages, document-level simplification in morphologically rich, low-resource languages such as Estonian remains largely unexplored. This study presents a comprehensive evaluation of five state-of-the-art multilingual large language models (LLMs) for document-level simplification in Estonian. Three prompting strategies are examined: single-pass generation, pipeline-based modular agents, and guideline-augmented pipelines. The evaluation framework integrates automatic metrics assessing readability, semantic preservation, and discourse coherence, alongside a structured manual annotation protocol. The findings indicate that \texttt{Gemini-2.0} and \texttt{LLaMA-3.3} produce outputs with near-native fluency and strong meaning preservation, whereas other models display notable grammatical and semantic limitations. This work contributes novel document-level coherence metrics, evidence-based prompting strategies, and publicly available resources for reproducibility.
\end{abstract}

\section{Introduction}

Text simplification aims to reduce linguistic complexity while preserving the original meaning of a text. This task improves access to information for language learners, individuals with cognitive impairments, and broader audiences seeking clarity \cite{alva-manchego-etal-2020-data}. Traditional research has focused on sentence-level simplification, involving transformations such as lexical substitution, syntactic restructuring, and sentence shortening \cite{zhang-lapata-2017-sentence}. These operations are vital for real-world use cases, including education, public communication, and assistive technologies \cite{readi-2020-tools, BARBU20155076}.

Document-level simplification presents additional challenges. Beyond sentence-internal complexity, it involves discourse coherence, anaphora resolution, and consistent structural transformations across paragraphs \cite{sun-etal-2021-document}. Sentence-by-sentence simplification can disrupt discourse structure and cohesion, potentially compromising comprehension \cite{vasquez-rodriguez-etal-2024-simple}.

This study addresses document-level simplification for Estonian, a morphologically rich and low-resource language. While large language models (LLMs) have demonstrated strong performance in sentence-level simplification for English, their document-level capabilities in low-resource settings remain insufficiently explored. To address this gap, five state-of-the-art multilingual LLMs are evaluated on Estonian Wikipedia articles using three prompting strategies.

This paper is guided by two research questions:

\begin{itemize}
\item \textbf{RQ1:} Which out-of-the-box large language models are most effective for document-level simplification in Estonian?
\item \textbf{RQ2:} Can the best-performing model be reliably identified using automatic evaluation metrics alone?
\end{itemize}

Contributions are threefold:
\begin{itemize}
\item Development of document-level prompting strategies for Estonian text simplification using multilingual LLMs;
\item Introduction of automatic coherence metrics based on sentence similarity and discourse marker preservation;
\item Design of a manual evaluation protocol that captures fluency, grammaticality, meaning preservation, and discourse-level coherence.
\end{itemize}

The remainder of this paper is structured as follows. Section~\ref{related_work} reviews prior work on document simplification. Section~\ref{llm} motivates the selection of large language models, describes the document-level simplification operations, and presents the prompting strategies used in our experiments. Section~\ref{setup} outlines the experimental setup, including model access, configuration, and evaluation data. Section~\ref{Results} details both the automatic and manual evaluation procedures. Finally, the paper concludes with a summary of findings and a brief overview of the public repository containing the code and resources necessary to reproduce our experiments.

\section{Related Work} \label{related_work}

Automatic text simplification (ATS) \cite{saggion2017automatic, Shardlow2014, espinosa-zaragoza-etal-2023-review} has evolved from rule-based approaches to neural architectures and large-scale pretrained models. Early systems relied on handcrafted rules for lexical substitution and syntactic transformations \cite{chandrasekar-etal-1996-motivations}, but scalability and domain adaptation posed challenges. Seq2Seq architectures improved fluency and structural variation, while reinforcement learning \cite{zhang-lapata-2017-sentence} and controllable models \cite{maddela-etal-2021-controllable} introduced optimization over multiple simplification objectives, such as grammaticality and simplicity.
Pretrained Transformer models like T5 \cite{10.5555/3455716.3455856} and GPT variants \cite{openai2023gpt4} enabled zero- and few-shot simplification, though concerns persist regarding cost, reproducibility, and closed-source APIs.

While sentence-level simplification has dominated research, document-level simplification remains comparatively underexplored. \citet{sun-etal-2021-document} introduced the D-Wikipedia dataset and proposed the D-SARI metric designed explicitly for evaluating document-level simplification, emphasizing the need to preserve discourse structure while simplifying. \citet{fang-etal-2025-progressive} proposed a progressive simplification method (ProgDS) that decomposes document simplification into hierarchical stages: discourse-level, topic-level, and lexical-level simplification, mimicking human editor workflows. Document-level simplification requires attention to coherence and cohesion \cite{siddharthan2006syntactic, vasquez-rodriguez-etal-2024-simple}. For example, \citet{siddharthan2006syntactic} showed that syntactic simplification must preserve anaphoric and conjunctive cohesive relations to maintain text coherence, while recent work by \citet{vasquez-rodriguez-etal-2024-simple} introduces SimDoc, a system that incorporates readability and coherence objectives during training, demonstrating the importance of multi-dimensional evaluation beyond simple lexical and syntactic measures.

Parallel to model development, corpora such as the PWKP dataset \cite{zhu-etal-2010-monolingual}, its later filtered version WikiSmall \cite{zhang-lapata-2017-sentence} along with TurkCorpus \cite{xu-etal-2016-optimizing} and Newsela \cite{xu-etal-2015-problems}, have supported supervised training and evaluation. However, simplification remains under-explored for low-resource languages. \citet{ryan-etal-2023-revisiting} introduced MultiSim, a multilingual benchmark covering 12 languages and over 1.7 million sentence pairs, demonstrating that multilingual training can enhance performance in non-English settings and enable cross-lingual transfer to low-resource languages. Despite this progress, Estonian is not among the languages included, highlighting a gap in current multilingual simplification resources. For Estonian specifically, existing work has focused on sentence-level transformations, including lexical substitution and template-based syntactic edits. \citet{barbu-muru-malva:2025:RANLP} introduced new Estonian datasets and fine-tuned LLaMA models, showing that large language models outperform NMT-based approaches in grammaticality, readability, and meaning preservation. The challenges of adapting simplification systems to morphologically rich, low-resource languages have been documented for other Baltic languages like Lithuanian \cite{mandravickaite-etal-2024-exploring}.

This study builds on previous work by focusing on document-level simplification for Estonian and introducing both automatic, discourse-aware evaluation and human annotation. It implements a four-agent simplification framework, inspired by the role-based prompting strategy proposed by \cite{fang-etal-2025-collaborative}.
This research addresses a critical gap in simplification research for morphologically rich, low-resource languages.

\section{Large Language Model Prompting Strategies for Document Simplification} \label{llm}

This section outlines the LLMs used in our experiments and the prompting strategies designed to elicit high-quality document-level simplifications. We begin by motivating our selection of both proprietary and open-access instruction-tuned LLMs. Next, we identify and describe the core set of document-level simplification operations that inform both evaluation and prompt design. Finally, we present the prompting workflows developed to enhance coherence, adequacy, and user-centered simplification in Estonian.

\subsection{Document-Level Simplification Operations}

Document-level simplification requires structural transformations that extend beyond isolated sentence rewriting. Core operations include \textit{sentence joining}, which fuses fragmented or related content to improve fluency; \textit{sentence reordering}, which reorganizes discourse for logical progression; and \textit{sentence splitting}, which breaks down syntactically complex sentences to ease comprehension. Other key operations are \textit{sentence deletion} which removes non-essential or redundant information, and \textit{sentence addition}, which introduces explanatory details to clarify difficult content. Finally, \textit{anaphora resolution} \cite{mitkov2002anaphora} plays a critical role in maintaining referential coherence, especially when prior operations disrupt the original structure. These transformations form the basis of our prompting strategies and help make document-level simplification more effective.

\subsection{Model Selection}

To investigate the effectiveness of LLMs for document-level text simplification in Estonian, five instruction-tuned models were selected based on their multilingual capabilities, accessibility, and performance in recent benchmarks. The models represent a spectrum of proprietary and open-access systems, varying in architecture size and geographic origin.

\begin{itemize}
    \item \textbf{GPT-4.1 (OpenAI)}: A proprietary model recognized for strong zero-shot and few-shot generalization across NLP tasks. Although its internal architecture remains undisclosed, it is widely regarded as a state-of-the-art language model \cite{openai2023gpt4}.

    \item \textbf{LLaMA-3.3-70B-Instruct (Meta)}: A 70-billion parameter model from Meta's LLaMA 3 series, released under an open-access license. The model is trained on diverse multilingual corpora and instruction-tuned, making it suitable for fine-tuned evaluation in research contexts \cite{touvron2023llamaopenefficientfoundation}.

    \item \textbf{Claude-3.5-Sonnet (Anthropic)}: A closed-source model designed for controllable and coherent text generation. It has demonstrated strong performance in reasoning and summarization tasks, making it relevant for simplification benchmarks \cite{anthropic_claude3.5_sonnet_2024}.

    \item \textbf{Gemini-2.0-flash-001 (Google DeepMind)}: Part of Google's Gemini family, this model is optimized for alignment, safety, and factuality. Despite being proprietary, it provides a valuable point of comparison due to its integration with widely used applications \cite{google_gemini2_2024}.

    \item \textbf{Qwen-2.5-72B-Instruct (Alibaba)}: A 72-billion-parameter multilingual model developed by Alibaba Cloud. As an open-access model with broad language coverage, Qwen-2.5-72B-Instruct provides a valuable benchmark for assessing document simplification performance in a low-resource language like Estonian \cite{qwen2025qwen25technicalreport}.

\end{itemize}

These models enable us to compare how well different systems perform document-level simplification, taking into account differences in openness, size, and language support; especially in the context of a low-resource language like Estonian.

\subsection{Prompting Strategies}

Three prompting strategies were used to help the models produce clear and readable document-level simplifications. Each one had a specific prompt tailored to the task. Examples are shown below, and all prompt templates can be found in the project repository (see Section~\ref{reproducibility}).

\paragraph{Single-Pass Prompting}
The instructions are provided within a single prompt, making the approach efficient but prone to generating outputs that closely resemble the original text, with only minimal simplification.

\begin{tcolorbox}[breakable,
  title={Prompt Example: Single-Pass Simplifier},
  label=prompt:single_pass,
  colback=gray!5, colframe=gray!40!black, fonttitle=\bfseries,
  left=6pt,right=6pt,top=6pt,bottom=6pt]
\textbf{Instruction:} \\
As a text simplification writer, your task is to simplify the given document. The simplified document should retain the original information. Elaborate complex terms or replace them with simpler alternatives. You may simplify the text via document-level simplification operations:

\begin{itemize}
  \item \textbf{Deleting Sentences:} Remove non-crucial or repetitive information.
  \item \textbf{Joining Sentences:} Merge adjacent fragments to improve fluency if deletion caused disfluency.
  \item \textbf{Splitting Sentences:} Divide long or complex sentences into simpler ones.
  \item \textbf{Reordering Sentences:} Improve logical flow by adjusting sentence order.
\end{itemize}

Apply these only when necessary. \\
\textbf{Output:} Provide only the simplified document in Estonian.
\end{tcolorbox}

\paragraph{Role-Based Agent Prompts}
Prompts are formulated to simulate the behavior of a domain-specific simplification agent (e.g., "You are a professional text simplifier working for Estonian public communication\ldots{}"). This approach aims to improve task alignment and output consistency by encouraging goal-oriented model behavior. Two distinct strategies were developed to investigate the effectiveness of role-based prompting in document-level text simplification.

\paragraph{Pipeline-Only}
The Pipeline-Only strategy employs a sequential workflow consisting of three specialized agents: the Terminology Interpreter (TI), the Structural Simplifier (SS), and the Anaphora Agent (AA). The output produced by each agent serves as the input for the subsequent one, forming a stepwise refinement process. The final simplified text is generated by the Anaphora Agent.

\begin{tcolorbox}[breakable,
  title={Prompt Example: Anaphora Agent},
  label=prompt:anaphora_agent,
  colback=gray!5, colframe=gray!40!black, fonttitle=\bfseries,
  left=6pt,right=6pt,top=6pt,bottom=6pt]
\textbf{Instruction:} \\
You are an anaphora agent reviewing a previously simplified Estonian document. Ensure all anaphors (e.g., pronouns and references) are coherent and correctly linked to their antecedents. Where necessary, elaborate or adjust unclear references. Remove redundant or repetitive content.

If everything is already correct, make no changes.

\textbf{Output:} Provide only the edited simplified document in Estonian.
\end{tcolorbox}

\paragraph{Pipeline-Guideline}
The Pipeline-Guideline (PG) strategy extends the basic pipeline by introducing an additional agent, the Project Director (PD), responsible for generating simplification guidelines. This agent operates at the outset of the process to establish clear objectives and stylistic constraints for the downstream agents. By providing these structured guidelines, the strategy aims to enhance the consistency, coherence, and goal alignment of the overall simplification workflow.

\begin{tcolorbox}[breakable,
  title={Prompt Example: Project Director},
  label=prompt:terminology_interpreter,
  colback=gray!5, colframe=gray!40!black, fonttitle=\bfseries,
  left=6pt,right=6pt,top=6pt,bottom=6pt]
\textbf{Instruction:} \\
You are a project director and your task is to create guidelines in Estonian for simplifying Estonian documents. The guidelines aim is to aid subsequent agents in the simplification process. The guideline should be set up as follows:

1. Executive Summary: A brief overview of the document's main points, arguments, and conclusions.

2. Terminology: A list of specialized terms, along with their definitions and contextual meanings. All uncommon words and specialized terms mentioned in the document should be explained.

3. Main Arguments and Evidence: An outline of the document's main arguments and the key evidence or examples that support these arguments.

4. Cultural and Social Context: Information about the cultural, historical, or social context mentioned in the document, to be appropriately addressed during simplification.

\textbf{Output:} Provide only the guidelines for Estonian document-level simplification.
\end{tcolorbox}

The prompt templates were manually crafted and iteratively refined through pilot testing to optimize clarity, brevity, and alignment with the simplification task.

\section{Experimental Setup} \label{setup}

\subsection{Evaluation Data}

A representative set of 15 articles was randomly selected from a filtered 4,000-article sample of Estonian Wikipedia. Articles were constrained to a length between 50 and 250 words to ensure sufficient linguistic variation while maintaining feasibility for human annotation. Reference simplifications were authored independently by a native Estonian linguist following structured annotation guidelines. These simplifications were created without exposure to model outputs to avoid bias in evaluation.

\subsection{Model Access and Configuration}

All large language models were accessed via the OpenRouter API.\footnote{\url{https://openrouter.ai}} Standard parameters were used across models, including a temperature of 0.7 and top-p of 0.9, unless otherwise noted. Token limits were adjusted according to each model's capabilities, with higher limits used for models such as Claude-3.5 that support extended context windows. Requests were submitted in OpenAI-compatible chat format using authenticated API calls.

\subsection{Prompt Execution Pipelines}

Three prompting strategies were implemented to support different levels of control over the simplification process: Single-Pass, Pipeline-Only, and Pipeline-Guideline. Each strategy involved a distinct series of interactions with the LLMs, ranging from a single task-oriented instruction to multi-agent workflows. Agent roles were defined using natural language instructions to emulate domain-specific simplification behavior. In the Pipeline-Guideline strategy, an initial guideline-generating agent was added to establish simplification objectives before subsequent content transformation.

\subsection{Software Environment}

All experiments were conducted in a Python 3.11 environment. Standard open-source libraries were used for API communication, text processing, and evaluation. These included packages for computing readability (e.g., FKGL), semantic similarity (e.g., BERTScore), and coherence (e.g., sentence embeddings). Sentence segmentation and discourse marker identification were conducted using Estonian-specific NLP tools. Evaluation and model interaction were performed on local hardware with GPU support.

\section{Results} \label{Results}

To assess the effectiveness of the evaluated models and prompting strategies, both automatic and manual evaluations were conducted.

\subsection{Automatic Evaluation}

Automatic evaluation was conducted using metrics that assess surface-level rewriting quality and discourse-level textual properties. The evaluation covers multiple LLMs' prompting strategies (Single-Pass Prompting, Pipeline-Only, and Pipeline-Guideline).

\paragraph{Surface-Level Metrics} The following metrics were used to quantify readability, semantic preservation, and lexical simplification:

\begin{itemize}
    \item \textbf{FKGL} \cite{kincaid1975derivation} estimates the U.S. school grade level required to understand a text, based on average sentence length and syllable count per word. Lower scores indicate simpler, more readable output. As a non-reference metric, FKGL is widely used in text simplification to assess surface-level readability, especially when gold-standard references are unavailable. Despite its simplicity and interpretability, FKGL can be gamed and does not capture meaning preservation or grammaticality. In this study, FKGL is used to complement semantic and structural metrics, and was computed using the \texttt{textstat} package.

    \item \textbf{BERTScore} \cite{Zhang2019BERTScoreET} measures semantic similarity between texts by comparing contextualized token embeddings from BERT. The measure is well-suited for evaluating document-level simplifications, where rephrasing and structural changes are common. Unlike surface-level metrics such as BLEU, BERT-S captures meaning preservation more effectively, aligning closely with human judgments. This study computed BERT-S using the official \texttt{bert\_score} package.\footnote{\url{https://github.com/Tiiiger/bert\_score}}

    \item \textbf{Document-SARI (D-SARI)} \cite{sun-etal-2021-document} extends the sentence-level SARI metric \cite{xu-etal-2016-optimizing} to evaluate simplification quality at the document level. It operates by computing sentence-wise SARI scores, assessing additions and deletions, keeping words between system output, source, and reference, and then aggregating these across documents. Unlike traditional n-gram metrics, D-SARI captures the simplification process by rewarding informative insertions and deletions while penalizing unnecessary retention.

\end{itemize}

\paragraph{Discourse-Level Metrics} To complement surface-level evaluation, two custom metrics were implemented to capture the simplified texts' discourse coherence and structural consistency. These metrics assess properties not reflected in traditional word or sentence-level scores.

\begin{itemize}
    \item \textbf{Coherence} was assessed by computing the average pairwise cosine similarity between adjacent sentence embeddings, using a multilingual Sentence-BERT model \cite{Reimers2019SentenceBERTSE}. This embedding-based approach quantifies the logical flow between consecutive sentences, with higher values indicating stronger local Coherence. We report both the mean similarity (\textit{coherence\_mu}) and its variability (\textit{coherence\_std}).

    \item \textbf{Discourse Marker Preservation} was measured as the proportion of discourse markers retained between the original and simplified versions. Markers were identified using a curated Estonian inventory based on established discourse relation taxonomies \cite{prasad-etal-2008-penn}. This metric reflects the degree to which discourse structure is preserved in the output. Both average preservation rate and standard deviation (\textit{discourse\_preservation}, \textit{discourse\_std}) are reported.
\end{itemize}

\paragraph{Automatic Evaluation Results}
Table~\ref{tab:auto-eval} reveals important trends across LLMs and prompting strategies. First, lower FKGL scores in the Single-Pass setting (SP) indicate higher textual simplicity; however, this often coincides with reduced BERT-S and D-SARI scores, highlighting a trade-off between simplicity and meaning preservation. In contrast, the Pipeline-Only (PO) and Pipeline-Guideline (PG) strategies yield higher BERT-S values, suggesting better retention of semantic content. D-SARI scores, which reward aligned addition, deletion, and retention operations, also tend to be higher for PO, indicating more controlled structural simplifications.

\begin{table*}[t]
\centering
\scriptsize
\setlength{\tabcolsep}{3pt}
\begin{tabular}{lccccccccccccccc}
\toprule
\textbf{Metric} & \multicolumn{3}{c}{Claude-3.5-Sonnet} & \multicolumn{3}{c}{Gemini-2.0-flash-001} & \multicolumn{3}{c}{GPT-4.1} & \multicolumn{3}{c}{LLaMA-3.3-70B-Instruct} & \multicolumn{3}{c}{Qwen-2.5-72B-Instruct} \\
& SP & PO & PG & SP & PO & PG & SP & PO & PG & SP & PO & PG & SP & PO & PG \\
\midrule
FKGL & 8.97 & 10.19 & 11.41 & 8.75 & \cellcolor{gray!20}9.75 & 9.41 & 10.29 & 9.99 & 10.75 & 9.65 & \cellcolor{gray!20}10.30 & 11.58 & 9.82 & \cellcolor{gray!20}9.79 & 10.65 \\
BERT-S & 0.888 & 0.876 & 0.863 & 0.899 & \cellcolor{gray!20}0.887 & 0.888 & 0.898 & 0.890 & 0.860 & 0.895 & \cellcolor{gray!20}0.883 & 0.882 & 0.886 & \cellcolor{gray!20}0.880 & 0.872 \\
D-SARI & 0.283 & 0.254 & 0.175 & 0.304 & \cellcolor{gray!20}0.274 & 0.269 & 0.227 & 0.242 & 0.020 & 0.310 & \cellcolor{gray!20}0.265 & 0.212 & 0.269 & \cellcolor{gray!20}0.271 & 0.179 \\
Coherence~$\mu$ & 0.394 & 0.344 & 0.404 & 0.405 & \cellcolor{gray!20}0.403 & 0.406 & 0.446 & 0.437 & 0.413 & 0.449 & \cellcolor{gray!20}0.430 & 0.429 & 0.401 & \cellcolor{gray!20}0.409 & 0.423 \\
Coherence~$\sigma$ & 0.086 & 0.087 & 0.079 & 0.100 & \cellcolor{gray!20}0.112 & 0.072 & 0.108 & 0.131 & 0.084 & 0.114 & \cellcolor{gray!20}0.133 & 0.125 & 0.115 & \cellcolor{gray!20}0.109 & 0.108 \\
Discourse~$\mu$ & 0.300 & 0.306 & 0.194 & 0.456 & \cellcolor{gray!20}0.400 & 0.400 & 0.494 & 0.433 & 0.367 & 0.433 & \cellcolor{gray!20}0.422 & 0.367 & 0.478 & \cellcolor{gray!20}0.478 & 0.328 \\
Discourse~$\sigma$ & 0.374 & 0.403 & 0.295 & 0.465 & \cellcolor{gray!20}0.436 & 0.436 & 0.470 & 0.462 & 0.446 & 0.462 & \cellcolor{gray!20}0.440 & 0.404 & 0.483 & \cellcolor{gray!20}0.445 & 0.394 \\
\bottomrule
\end{tabular}
\caption{Automatic evaluation results for each LLM and prompting strategy. SP = Single-Pass Prompting, PO = Pipeline-Only, PG = Pipeline-Guideline. Metrics: FKGL (readability), BERT-S (semantic similarity), D-SARI (simplification operations), Coherence (cosine similarity of adjacent sentences), Discourse (marker preservation). The strategies that were manually evaluated are indicated in gray for clarity.}
\label{tab:auto-eval}
\end{table*}

Discourse-level metrics provide deeper insights. Coherence is measured by pairwise cosine similarity of adjacent sentence embeddings, while Discourse evaluates the preservation of discourse markers. Both metrics tend to show a consistent advantage for the PG strategy. Notably, \texttt{GPT-4.1} and \texttt{LLaMA-3.3} achieve the highest coherence scores (0.413 and 0.429), while \texttt{Gemini-2.0} and \texttt{Qwen-2.5} maintain strong performance across both discourse and coherence dimensions. In contrast, \texttt{Claude-3.5} shows lower discourse preservation under PG (0.194), suggesting difficulties maintaining explicit structural cues despite improved coherence.

\texttt{GPT-4.1} performs strongly in SP and PO settings, especially in terms of semantic coherence and discourse preservation. However, it underperforms in the PG setting, where structural fidelity and simplification operations degrade significantly. \texttt{Qwen-2.5}, while consistent in discourse-related scores, lags slightly in semantic similarity and grammaticality compared to top-performing models like \texttt{Gemini-2.0} and \texttt{LLaMA-3.3}.

These results show that the prompting strategy can significantly affect output quality, sometimes more than the LLM architecture.

\begin{figure}[ht]
    \centering
    \includegraphics[width=0.5\textwidth]{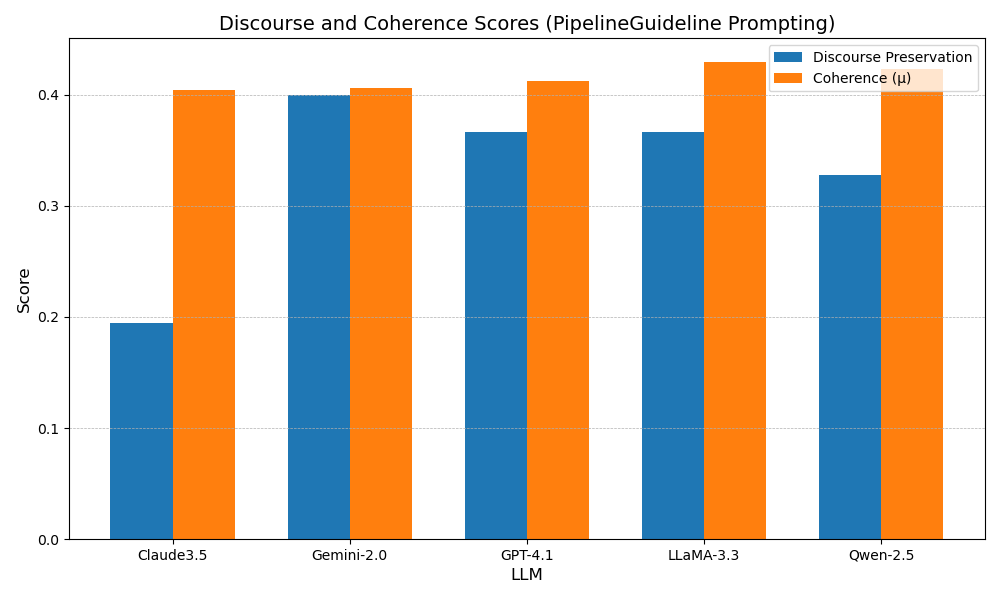}
    \caption{Discourse preservation and coherence scores across LLMs using the Pipeline-Guideline prompting strategy.}
    \label{fig:discourse-coherence}
\end{figure}

Figure~\ref{fig:discourse-coherence} complements Table~\ref{tab:auto-eval} by providing a visual comparison of discourse-level performance under the Pipeline-Guideline (PG) strategy. Notably, \texttt{LLaMA-3.3} and \texttt{GPT-4.1} achieve the highest coherence scores, suggesting strong logical flow and sentence connectivity in their simplified outputs. In terms of discourse marker preservation, \texttt{Gemini-2.0} outperforms the other models, reflecting its strength in retaining explicit textual cues such as connectives and transitions.

The contrast between coherence and discourse scores highlights distinct model-specific tendencies. For instance, \texttt{Claude-3.5} demonstrates relatively low discourse marker preservation despite achieving moderate coherence, indicating that while it produces fluent text, it often omits explicit structural anchors. In contrast, \texttt{Gemini-2.0} maintains a strong balance between semantic flow and discourse structure, performing consistently across both dimensions. \texttt{Qwen-2.5} shows a different pattern: it preserves surface-level discourse markers reasonably well but struggles with coherence, suggesting difficulties in maintaining fluid progression between sentences.

These findings underscore the importance of evaluating implicit and explicit discourse features in simplification outputs. While coherence captures the internal semantic continuity of the text, discourse marker preservation offers a complementary perspective on structural fidelity.

It also motivates the manual evaluation that follows.

\subsection{Manual Evaluation}

Based on the automatic evaluation results, we selected three top-performing models for manual assessment: \texttt{Gemini-2.0}, \texttt{Qwen-2.5}, and \texttt{LLaMA-3.3}. These models exhibited the highest overall performance in terms of fluency, meaning preservation, and readability.

After an initial qualitative analysis of outputs from all prompting strategies, the manual evaluation focused exclusively on the Pipeline-Only approach. This method consistently yielded the most coherent and structurally sound document-level simplifications. In contrast, though effective at breaking down complex content, the Pipeline-Guideline strategy often led to over-segmentation and overly verbose explanatory additions.

\paragraph{Manual Evaluation Workflow}
Manual evaluation aimed to assess the quality of the simplified Estonian texts along three core dimensions:

\begin{itemize}
    \item \textbf{Grammaticality}: This dimension evaluates whether the simplified text uses correct case forms, verb conjugations, word order, and punctuation. It reflects the overall grammatical integrity and naturalness of the output.

    \item \textbf{Meaning Preservation}: This criterion assesses how accurately the simplified version conveys the essential information and intent of the original text. It ensures that no critical details are omitted, distorted, or replaced by unwarranted additions that alter the intended meaning.

    \item \textbf{Simplicity}: This dimension examines how much lexical and syntactic complexity has been reduced. Simplicity is achieved through more precise vocabulary, shorter and more manageable sentences, and improved structural organization. Annotators also considered whether document-level simplification operations were effectively applied.
\end{itemize}

Two native Estonian speakers with expertise in linguistics conducted the annotation. Before beginning the main evaluation, the annotators held a calibration session to ensure consistency. This meeting clarified the evaluation criteria, resolved ambiguities in interpretation, and aligned their use of the 5-point Likert scale. The full annotation guidelines used during this process are publicly available in our repository (see Section~\ref{reproducibility}).

\paragraph{Inter-Annotator Agreement}
Both annotators independently rated the same document simplified by each of the three evaluated models to assess reliability. Ratings were collected for grammaticality, meaning preservation, and simplicity using a 5-point Likert scale. The inter-annotator agreement was measured using \textit{quadratic-weighted Cohen's $\kappa$} \cite{cohen1968weighted}, which accounts for ordinal rating differences.

The agreement was substantial across all three dimensions. The highest agreement was observed for \textit{Meaning Preservation} ($\kappa = 0.73$), followed by \textit{Grammaticality} and \textit{Simplicity} (both at $\kappa = 0.67$). These values confirm the consistency of the annotation protocol and support the reliability of the following evaluation results.

Following this reliability phase, the annotators independently evaluated four additional articles each, thus producing nine annotated documents.

\paragraph{Manual Evaluation Results}
Table~\ref{tab:manual-eval} shows clear differences in how the three evaluated models performed. \texttt{Gemini-2.0} scored the highest in all areas. \texttt{LLaMA-3.3} was close behind, while \texttt{Qwen-2.5} performed noticeably worse. Figure~\ref{fig:simplification-example} provides a qualitative comparison of the original input and the simplifications generated by the top-performing models.

\begin{figure}[ht]
\centering

\begin{tcolorbox}[title=Original Text]
Ajupõletik ehk entsefaliit (ladina ja inglise keeles encephalitis) on
põletikuline ajuaine haigus, mida põhjustab haigusetekitaja otsene
tungimine ajukoesse või immuunsüsteemi ülitundlikkusreaktsioon
(autoimmuunne entsefaliit) viiruse või võõrvaigu suhtes.
\end{tcolorbox}

\begin{tcolorbox}[title=Gemini-2.0]
Aju põletik ehk entsefaliit on haigus, kus aju kude põletikku läheb.
Seda võivad põhjustada haigustekitajad või immuunsüsteemi liigne
reaktsioon.
\end{tcolorbox}

\begin{tcolorbox}[title=LLaMA-3.3]
Aju põletik (encefaliit) on ajuhaigus, mis tekib, kui haigusetekitaja
tungib ajju või kui immuunsüsteem reageerib liigselt viiruse või
võõrakehale.
\end{tcolorbox}

\begin{tcolorbox}[title=Qwen-2.5]
Ajupõletik on haigus, mis põhjustab aju survekahju. Selle põhjustavad kas mikroobid, näiteks viirused, või immuunsüsteemi üleloomlik reageerimine.
\end{tcolorbox}

\caption{Example of text simplification outputs from different models.}
\label{fig:simplification-example}

\end{figure}

\begin{table*}[t]
\centering
\scriptsize
\setlength{\tabcolsep}{6pt}
\begin{tabular}{lcccc}
\toprule
\textbf{Model} & \textbf{Grammaticality} & \textbf{Meaning Preservation} & \textbf{Simplicity} & \textbf{Average} \\
\midrule
Qwen-2.5-72B-Instruct& 1.38 & 2.25 & 1.63 & 1.75 \\
Gemini-2.0-flash-001  & 4.88 & 4.75 & 4.75 & 4.79 \\
LLaMA-3.3-70B-Instruct& 4.00 & 4.25 & 4.25 & 4.17 \\
\bottomrule
\end{tabular}
\caption{Manual evaluation results for three top-performing models. Scores are averaged across evaluated articles using a 5-point Likert scale (1 = Very Poor, 5 = Excellent).}
\label{tab:manual-eval}
\end{table*}

\texttt{Gemini-2.0} consistently delivered fluent, coherent, and well-structured simplifications annotators found much easier to understand than the originals. In several cases, it earned perfect scores across all evaluation dimensions. While it occasionally made minor grammatical errors, such as incorrect case endings, these did not interfere with readability or meaning. The model preserved essential information and avoided distortions. Although it sometimes chose slightly off-target words, these issues were rare and did not impact the overall quality. As shown in Figure~\ref{fig:simplification-example}, the model substantially shortens the original sentence by removing less essential details (e.g., references to other languages) and splitting the content into two sentences. These results suggest that \texttt{Gemini-2.0} handles Estonian document-level simplification effectively.

Overall, \texttt{LLaMA-3.3} performed well. It mostly produced grammatical, well-organized texts with only small omissions of information. Annotators consistently rated its outputs as easier to read than the original versions. While it clarified the text, it often retained unnecessary complexity, such as unnecessary elaborations. In addition, the model introduced slight hallucinations and grammatical errors. Figure~\ref{fig:simplification-example} illustrates that the model introduces a lexical error by changing \textit{entsefaliit} to \textit{encefaliit}, which appears to be a mixture of the Estonian term \textit{entsefaliit} and its Latin/English counterpart \textit{encephalitis}.

In contrast, \texttt{Qwen-2.5} performed poorly across all evaluation dimensions. It often produced texts with grammatical errors, including incorrect case endings, fragmented sentences, and frequent misspellings. Annotators also observed significant meaning distortions, such as hallucinated content and omissions of essential information. While the model attempted various simplification strategies, its lack of grammatical control and weak alignment with the source text made the outputs difficult to follow. Figure~\ref{fig:simplification-example} also highlights lexical errors in the model output, including the non-existent words \textit{üleloomik} and \textit{survekahju}, which also diverge the meaning of the text. In general, the simplified versions were more confusing than the originals. The model has limited Estonian support and underscores the challenges of adapting multilingual models to morphologically rich, low-resource languages.

The manual evaluation confirms that model performance in document-level simplification is highly variable. While models like \texttt{Gemini-2.0} and \texttt{LLaMA-3.3} produce near-native fluency and maintain semantic fidelity, others such as \texttt{Qwen-2.5} exhibit critical deficiencies. These findings underscore the importance of careful evaluation when deploying LLMs for complex language processing tasks in low-resource settings.

\paragraph{Comparison of Manual and Automatic Evaluation}
The results of the manual evaluation align closely with some automatic metrics for the Pipeline-Only strategy, particularly for the \texttt{Gemini-2.0} and \texttt{LLaMA-3.3} models. \texttt{Gemini-2.0} received the highest average manual score (4.79), consistent with its strong performance in BERTScore (0.887) and D-SARI (0.274), indicating high semantic fidelity and effective simplification operations. \texttt{LLaMA-3.3} ranked second in manual evaluation (4.17) and also ranked second in automatic evaluation (BERTScore 0.883, D-SARI 0.265), though annotators noted a slightly conservative simplification style. In contrast, \texttt{Qwen-2.5}, which scored the lowest in manual evaluation (1.75), also showed weaker automatic performance under PO, with the lowest BERTScore (0.880) and only moderate D-SARI (0.271).

\section{Conclusion} \label{conclusion}

This study investigated two research questions: \textbf{RQ1:} which out-of-the-box large language models are most effective for document-level simplification in Estonian, and \textbf{RQ2:} whether automatic metrics alone can reliably identify the best-performing model.

Across five instruction-tuned LLMs and three prompting strategies, the results show that \texttt{Gemini-2.0} and \texttt{LLaMA-3.3} are the most effective models for document-level simplification in Estonian. Both produce fluent outputs with strong meaning preservation, and their performance remains robust across evaluation dimensions. By contrast, \texttt{Qwen-2.5} frequently introduced grammatical errors and meaning distortions, highlighting that multilingual coverage does not guarantee reliable performance in morphologically rich, low-resource settings.

For \textbf{RQ2}, automatic metrics (FKGL, BERTScore, D-SARI) provided useful signals for broad model comparisons, and the proposed discourse-aware measures (sentence-embedding coherence and discourse marker preservation) offered complementary evidence about document structure. However, the alignment between automatic metrics and human judgment was only partial. Manual evaluation was necessary to capture qualitative distinctions that matter at the document level, including local coherence, referential clarity, and the usability of the simplified text as a whole. In addition, the qualitative inspection that motivated focusing on the Pipeline-Only strategy reinforces that prompt design decisions for document-level simplification benefit from human-in-the-loop assessment.

Overall, the findings demonstrate that multilingual LLMs can support document-level simplification in Estonian, but achieving reliable quality requires careful choices in both prompting strategy and evaluation methodology. All resources needed to reproduce the experiments are made publicly available (Section~\ref{reproducibility}).

\section{Reproducibility} \label{reproducibility}

To ensure transparency and enable reproducibility, we share the full codebase, evaluation scripts, prompt templates, and model outputs used in our experiments. The repository includes:

\begin{itemize}
    \item Original Estonian Wikipedia articles and manually simplified references;

    \item LLM outputs for all prompting strategies:
    \texttt{SinglePass} (corresponds to \textbf{Single-Pass Prompting} in the paper),
    \texttt{PipelineOnly} (corresponds to \textbf{Pipeline-Only}),
    and \texttt{PipelineGuideline} (corresponds to \textbf{Pipeline-Guideline});

    \item Scripts for computing both sentence-level (FKGL, D-SARI, BERTScore) and document-level (discourse preservation, coherence) evaluation metrics;

    \item Prompt templates and agent definitions used in the simplification workflows.
\end{itemize}

The repository is publicly accessible in GitHub:
\begin{center}
\texttt{\small \url{https://github.com/meerimuru/ETDocSimpLREC2026}}
\end{center}

All results reported in the paper can be reproduced using the provided scripts. The only external dependency is access to the evaluated LLMs via the OpenRouter API.

\section{Limitations}
Several limitations should be considered when interpreting these findings. First, the manual evaluation was conducted on a small set of documents, which constrains the statistical strength of conclusions about fine-grained differences between models and prompting strategies. Second, the evaluation data consists of Estonian Wikipedia articles; while this enables controlled experimentation, it may not reflect the linguistic variability and pragmatic demands of other genres (e.g., public service communication, health information, or educational materials). Third, our models were evaluated in an out-of-the-box setting via a shared API interface; this design supports comparability, but it does not examine whether Estonian-specific adaptation (e.g., fine-tuning, terminology constraints, or decoding control) could substantially change the ranking of open-access models. Finally, the proposed discourse-aware automatic metrics capture only a subset of document-level quality: they quantify local semantic continuity and the preservation of explicit discourse markers, but they do not fully model global discourse structure, factual consistency, or the appropriateness of content additions and deletions.

\section{Acknowledgements}
This work was supported by the Estonian Research Council grant PRG2006.

\renewcommand{\refname}{Bibliographical References}

\bibliographystyle{lrec2026-natbib}
\bibliography{document_simplification}

\end{document}